\documentclass{article}

    \PassOptionsToPackage{compress}{natbib}

  \usepackage[preprint]{neurips_2019}

\usepackage[utf8]{inputenc} 
\usepackage[T1]{fontenc}    
\usepackage{hyperref}       
\usepackage{url}            
\usepackage{booktabs}       
\usepackage{amsfonts}       
\usepackage{nicefrac}       
\usepackage{microtype}      
\usepackage{graphicx}
\usepackage{subcaption}
\usepackage[dvipsnames]{xcolor}
\usepackage{wrapfig}
\usepackage{amsthm,amssymb,amsmath}

\usepackage{etoolbox}
\usepackage[ruled]{algorithm2e}
\usepackage{algpseudocode}
\usepackage{multirow}
\usepackage{hhline}

 \usepackage{ucs}
 \usepackage{comment}
 \usepackage{fancybox}
 \usepackage{pbox}
 \usepackage{tcolorbox}
 \usepackage{capt-of}
 \usepackage{bm}
 \usepackage{array}
 \usepackage{multirow}
 \usepackage{tikz}

\newcommand\wi[1]{$\circ$}
\newcommand\bu[1]{$\bullet$}
\newcommand\ot[1]{$\star$}
\newcommand\spa[1]{$\spadesuit$}
\newcommand\dn[1]{.}

\definecolor{myred}{rgb}{0.8,0,0}
\definecolor{mygreen}{rgb}{0,0.6,0}
\definecolor{myblue}{rgb}{0,0,0.7}

\newcounter{cbox} 
\newcommand{\cbox}{\arabic{cbox}}

\newcounter{cmes} 
\newcommand{\cmes}{\arabic{cmes}}

\makeatletter
\newcounter{algorithmbis}
\renewcommand{\thealgorithmbis}{\thesection.\arabic{algorithmbis}}
\def\algorithmbis{\@ifnextchar[{\@algorithmbisa}{\@algorithmbisb}}
\def\@algorithmbisa[#1]{%
  \refstepcounter{algorithmbis}
  \trivlist
  \leftmargin\z@
  \itemindent\z@
  \labelsep\z@
  \item[\parbox{\linewidth}{%
    \hrule
    \hrule
    \noindent\strut\textbf{Algorithm \thealgorithmbis} #1
    \hrule
  }]\hfil\vskip0em%
}
\def\@algorithmbisb{\@algorithmbisa[]}

\makeatother

\definecolor{myred}{rgb}{0.8,0,0}
\definecolor{mygreen}{rgb}{0,0.6,0}
\definecolor{myblue}{rgb}{0,0,0.7}

\allowdisplaybreaks 

\usepackage[usestackEOL]{stackengine}

\newcommand{\lstm}{{LSTM}\xspace}
\newcommand{\bubblesort}{{\sc bubblesort}\xspace}
\newcommand{\bubble}{{\sc bubble}\xspace}
\newcommand{\reset}{{\sc reset}\xspace}
\newcommand{\mcts}{{MCTS}\xspace}
\newcommand{\lstms}{{LSTM}s\xspace}
\newcommand{\stopnpi}{{\sc stop}\xspace}
\newcommand{\anpi}{AlphaNPI\xspace}
\newcommand{\npi}{{NPI}\xspace}
\newcommand{\npis}{{NPI}s\xspace}
\DeclareMathOperator*{\argmax}{argmax}

\newcounter{auxFootnote}

\title{Learning Compositional Neural Programs \\ with Recursive Tree Search and Planning}
\author{Thomas Pierrot\\
    InstaDeep\\
    \texttt{t.pierrot@instadeep.com}\\
    \And Guillaume Ligner\\
    InstaDeep\\
    \texttt{g.ligner@instadeep.com}\\
    \And Scott Reed\\
    DeepMind\\
    \texttt{reedscot@google.com}\\
    \AND Olivier Sigaud\\
    Sorbonne Universite\\
    \texttt{olivier.sigaud@upmc.fr}\\
    \And Nicolas Perrin\\
    Sorbonne Universite\\
    \texttt{perrin@isir.upmc.fr}\\
    \And Alexandre Laterre\\
    InstaDeep\\
    \texttt{a.laterre@instadeep.com}\\
    \AND David Kas\\
    InstaDeep\\
    \texttt{d.kas@instadeep.com}\\
    \And Karim Beguir\\
    InstaDeep\\
    \texttt{kb@instadeep.com}\\
    \And Nando de Freitas\\
    DeepMind\\
    \texttt{nandodefreitas@google.com}\\
}

\begin{document}

\maketitle

\begin{abstract}
We propose a novel reinforcement learning algorithm, \anpi, that incorporates the strengths of Neural Programmer-Interpreters (\npi) and AlphaZero. 
\npi contributes structural biases in the form of modularity, hierarchy and recursion, which are helpful to reduce sample complexity, improve generalization and increase interpretability. AlphaZero contributes powerful neural network guided search algorithms, which we augment with recursion. 
\anpi only assumes a hierarchical program specification with sparse rewards: 1 when the program execution satisfies the specification, and 0 otherwise. Using this specification, AlphaNPI is able to train NPI models effectively with RL for the first time, completely eliminating the need for strong supervision in the form of execution traces.
The experiments show that \anpi can sort as well as previous strongly supervised \npi variants.  The \anpi agent is also trained on a Tower of Hanoi puzzle with two disks and is shown to generalize to puzzles with an arbitrary number of disks. 
\end{abstract}

\section{Introduction}
Learning a wide variety of skills, which can be reused and repurposed to learn more complex skills or to solve new problems, is one of the central challenges of AI. As argued in \cite{bengio2019meta-transfer}, beyond achieving good generalization when both the training and test data come from the same distribution, we want knowledge acquired in one setting to transfer to other settings with different but possibly related distributions. 

Modularity is a powerful inductive bias for achieving this goal with neural networks \citep{parascandolo2018learning,bengio2019meta-transfer}. Here, we focus on a particular modular representation known as Neural Programmer-Interpreters (\npi) \citep{npi}. The \npi architecture consists of a library of learned program embeddings that can be recomposed to solve different tasks, 
a core recurrent neural network that learns to interpret arbitrary programs,
 and domain-specific encoders for different environments. \npi achieves impressive multi-task results, with strong improvements in generalization and reductions in sample complexity. While fixing the interpreter module, \cite{npi} also showed that \npi can learn new programs by re-using existing ones. 

The \npi architecture can also learn recursive programs. In particular, \cite{recursivenpi} demonstrates that it is possible to take advantage of recursion to obtain theoretical guarantees on the generalization behaviour of recursive \npis. Recursive \npis are thus amenable to verification and easy interpretation. 

The \npi approach at first appears to be very general because as noted in \citep{npi}, programs appear in many guises in AI; for example, as image transformations, as structured control policies, as classical algorithms, and as symbolic relations. However, \npi suffers from one important limitation: It requires supervised training form execution traces. This is a much stronger demand for supervision than input-output pairs. Thus the practical interest has been limited.

Some works have attempted to relax this strong supervision assumption.  \cite{li2016neural} and \cite{fox2018parametrized} train variations of \npi using mostly low-level demonstration trajectories but still require a few full execution traces. 
Indeed,  \cite{fox2018parametrized} states ``Our results suggest that adding weakly supervised demonstrations to the training set can improve performance at the task, but only when the strongly supervised demonstrations already get decent performance''.  

\cite{xiao2018improving} incorporate combinatorial abstraction techniques from functional programming into \npi. They report no difficulties when learning using strong supervision, but substantial difficulties when attempting to learn \npi models with curricula and REINFORCE. In fact, this policy gradient reinforcement learning (RL) algorithm fails to learn simple recursive \npis, attesting to the difficulty of applying RL to learn \npi models. 

This paper demonstrates how to train \npi models effectively with RL for the first time.
We remove the need for execution traces in exchange for a specification of programs and associated correctness tests on whether each program has completed successfully.
This allows us to train the agent by telling it \emph{what} needs to be done, instead of \emph{how} it should be done.
In other words, we show it is possible to overcome the need for strong supervision by replacing execution traces with a library of programs we want to learn and corresponding tests that assess whether a program has executed correctly. 

The user specifying to the agent what to do, and not how to do it is reminiscent of programmable agents \cite{denil2017programmable} and declarative vs imperative programming.
In our case, the user may also define a hierarchy in the program specification indicating which programs can be called by another. 

The RL problem at-hand has a combinatorial nature, making it exceptionally hard to solve. Fortunately, we have witnessed significant progress in this area with the recent success of AlphaZero \citep{silver2017mastering} in the game of Go. In the single-agent setting, \cite{rankedrewards} have demonstrated the power of AlphaZero when solving combinatorial bin packing problems. 

In this work, we reformulate the original \npi as an actor-critic network and endow the search process of AlphaZero with the ability to handle hierarchy and recursion. These modifications, in addition to other more subtle changes detailed in the paper and appendices, enable us to construct a powerful RL agent, named \anpi, that is able to train NPI models by RL\footnote{The code is available at \url{https://github.com/instadeepai/AlphaNPI}}. 

\anpi is shown to match the performance of strongly supervised versions of \npi in the experiments. The experiments also shed light on the issue of deploying neural network RL policies. We find that agents that harness Monte Carlo tree search (MCTS) planning at test time are much  more effective than plain neural network policies. 


\section{Problem statement and definitions}

We consider an agent interacting with an environment, choosing actions $a$ and making observations $e$. An example of this is bubble sort, where the environment is represented as a list of numbers, and the initial actions are one-step pointer moves and element swaps.  We call this initial set of actions \emph{atomic actions}.
As training progresses, the agent learns to profit from atomic actions to acquire higher-level programs. Once a program is learned, it is incorporated into the set of available actions. For example, in bubble sort, the agent may learn the program \reset, which moves all pointers to the beginning of the list, and subsequently the agent may harness the program \reset as an action. 

In our approach, a program has \emph{pre-conditions} and \emph{post-conditions}, which are tests on the environment state. All pre-conditions must be satisfied before execution. A program executes correctly if its post-conditions are verified. For example, the pre-condition for bubble sort is that both pointers are positioned at the beginning of the list. The post-condition is a test indicating whether the list is sorted upon termination. A program terminates when it calls the atomic action \stopnpi, which is assumed to be available in all environments. A \emph{level} is associated with each program, enabling us to define a hierarchy: Atomic actions have level 0 and any other program has a positive level. In our work, a program can only call lower-level programs or itself.

We formulate learning a hierarchical library of programs as a multi-task RL problem. In this setting, each task corresponds to learning a single program.  The action space consists of atomic actions and learned programs. The reward signal is $1$ if a program executes correctly, and $0$ otherwise. The agent's goal is to maximize its expected reward over all the tasks. In other words, it has to learn all the programs in the input specification.


\section{AlphaNPI}

\begin{figure}[ht!]
\includegraphics[width=14cm]{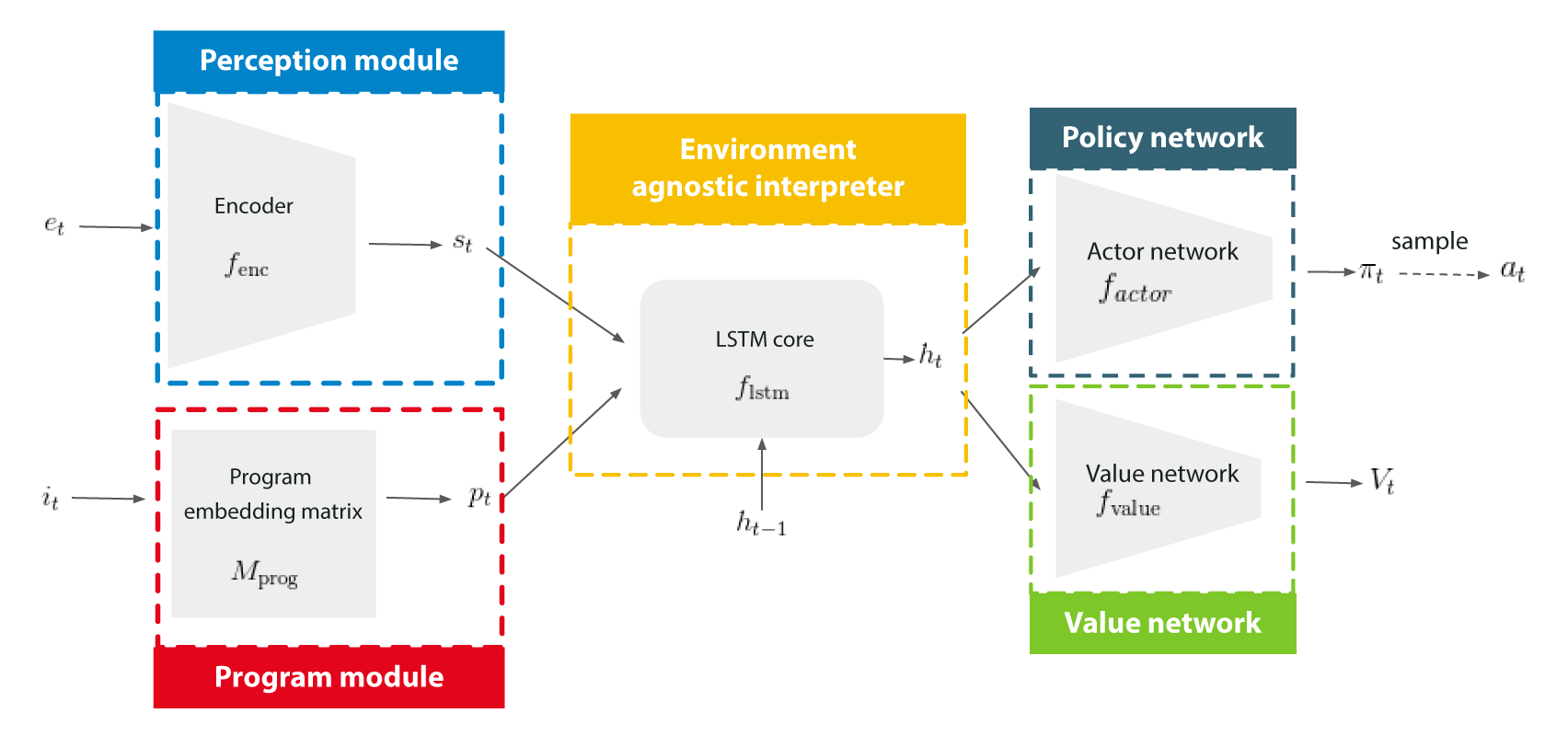}
\caption{\small \anpi modular neural network architecture.}
\label{fig:npi_architecture}
\end{figure}

Our proposed agent, \anpi, augments the \npi architecture of \cite{npi} to construct a recursive compositional neural network policy and a value function estimator, as illustrated in Figure~\ref{fig:npi_architecture}. It also extends the \mcts procedure of \cite{silver2017mastering} to enable recursion.
The \anpi network architecture consists of five modules: 
State (or observation) encoders, a program embedding matrix, an \lstm \citep{lstm} interpreter, a policy (actor) network and a value network. Some of these modules are universal, and some are task-dependent. The architecture is consequently a natural fit for multi-task learning.

Programs are represented as vector embeddings $p$ indexed by $i$ in a library. As usual, we use an embedding matrix for this.  The observation encoder produces a vector of features $s$. The universal \lstm core interprets and executes arbitrary programs while conditioning on these features and its internal memory state $h$. 
The policy network converts the \lstm output to a vector of probabilities $\pi$ over the action space, while the policy network uses this output to estimate the value function $V$. The architecture is summarized by the following equations:
\begin{equation}
        s_t = f_{\text{enc}}(e_t),\;
        p_{t} = M_{\text{prog}}[i_{t}, :],\;
        h_{t} = f_{\text{lstm}}(s_t, p_t, h_{t-1}),\;
        \pi_t = f_{\text{actor}}(h_t),\;
        V_t =  f_{\text{value}}(h_t).\;
\end{equation}
The neural nets have parameters, but we omit them in our notation to simplify the presentation.
These parameters and the program embeddings are learned simultaneously during training by RL.

When this \anpi network executes a program, it can either call a learned sub-program or itself (recursively), or perform an atomic action. When the atomic action is \stopnpi, the program terminates and control is returned to the calling program using a stack. When a sub-program is called, the stack depth increases and the \lstm memory state $h$ is set to a vector of zeroes. This turns out to be very important for verifying the model \citep{recursivenpi}.

\begin{figure}[t!]
\includegraphics[width=14cm]{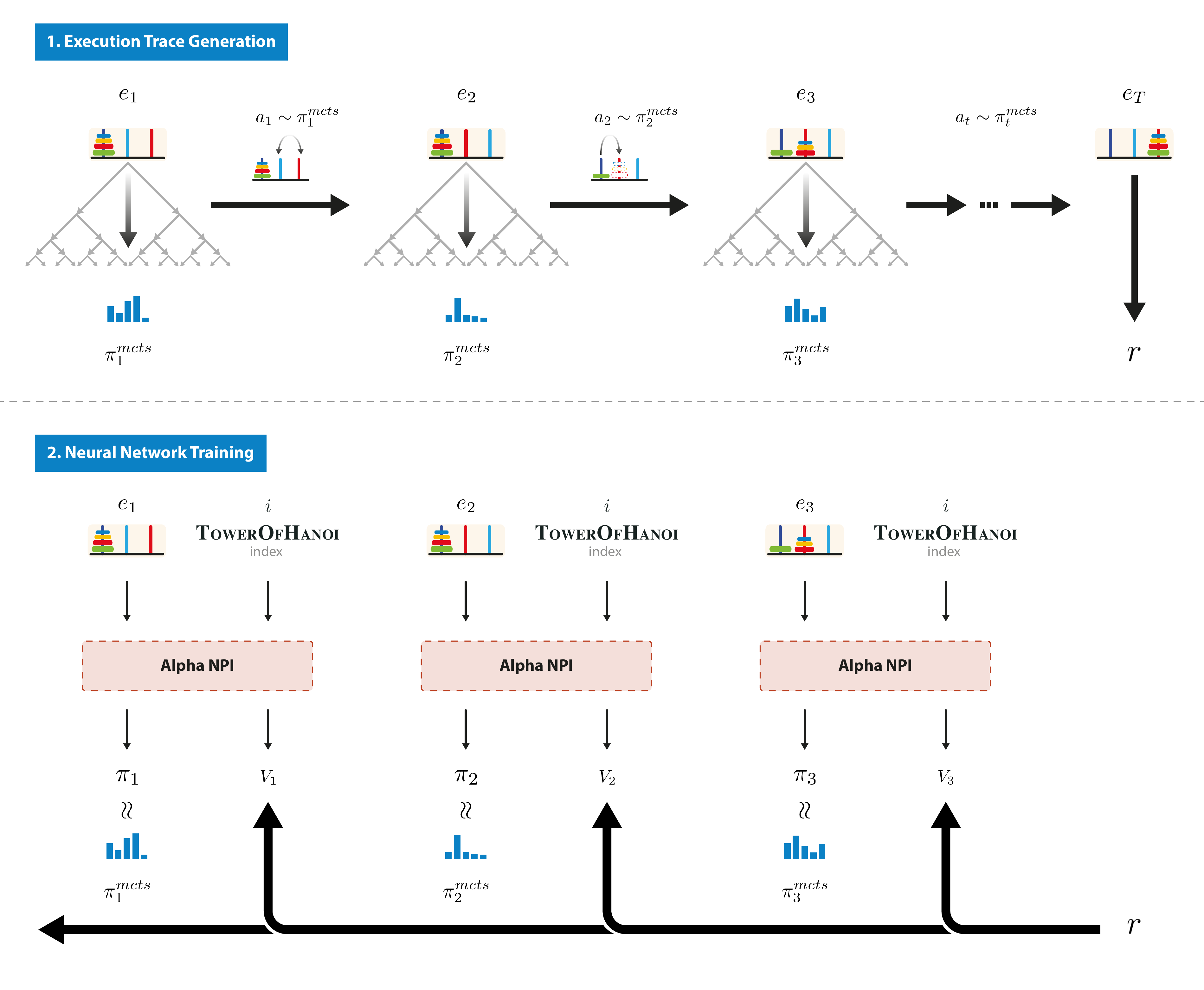}
\caption{\small\textbf{Execution trace generation with \anpi to solve the Tower of Hanoi puzzle.} 
\textbf{1}. To execute the $i$-th program, {\sc TowerOfHanoi}, AlphaNPI generates an execution trace $(a_1, \ldots, a_T)$, with observations $(e_1, \ldots, e_T)$ produced by the environment and actions $a_t \sim \pi_t^{mcts}$ produced by MCTS using the latest neural net, see Figure \ref{fig:npi_tree}. 
When the action \stopnpi is chosen, the program's post-conditions are evaluated to compute the final reward $r$.
The tuples $(e_t, i, h_t, \pi_t^{mcts}, r)$ are stored in a replay buffer.
\textbf{2}. The neural network parameters are updated to maximise the similarity of its policy vector output $\pi$ to the search probabilities $\pi^{mcts}$, and to minimise the error between the predicted value $V$ and the final reward $r$. To train the neural network, shown in Figure~\ref{fig:npi_architecture}, we use the data in the replay buffer.
}
\label{fig:npi_trace_generation}
\end{figure}

\begin{figure}[t!]
\includegraphics[width=14cm]{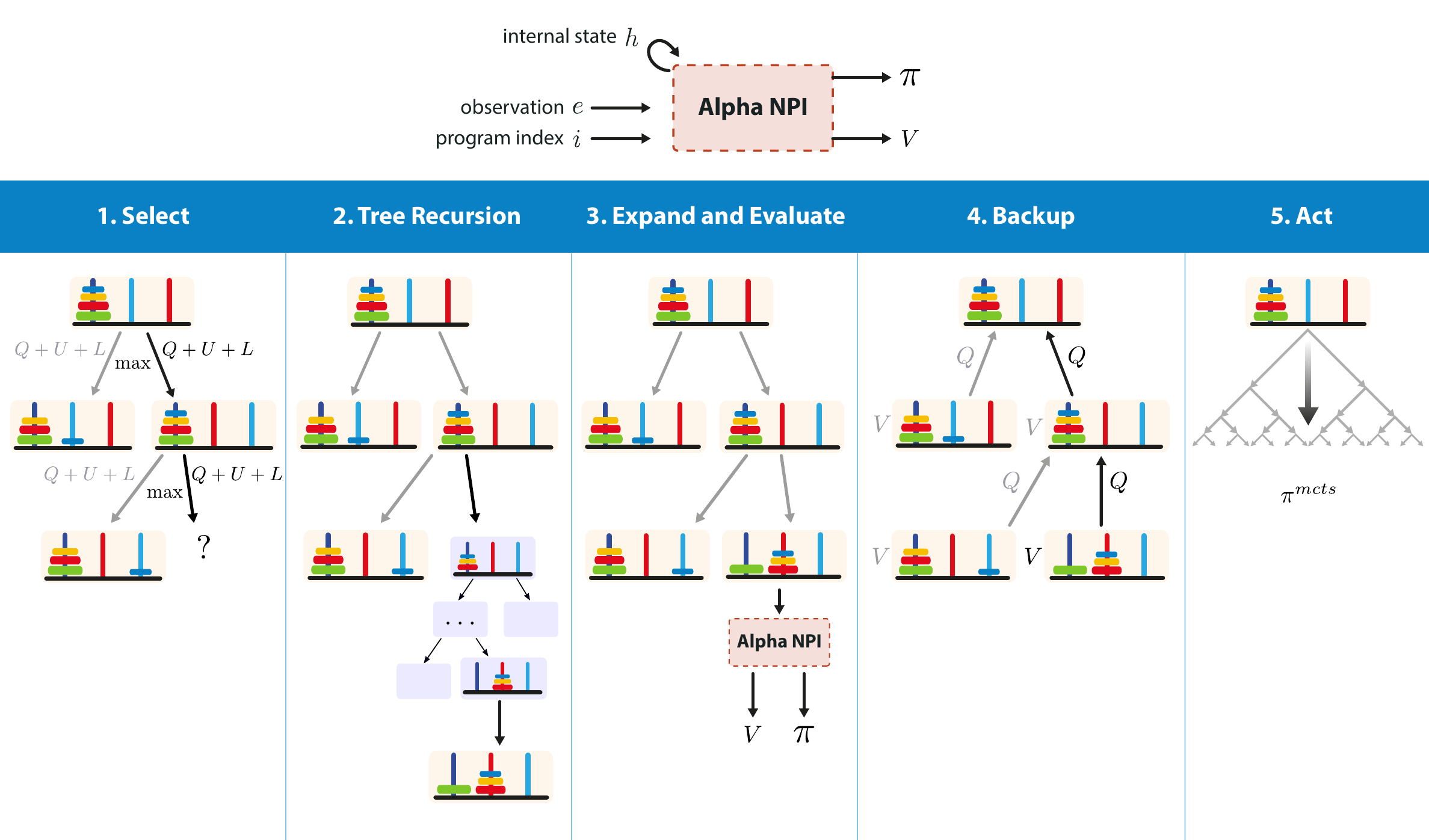}
\caption{\small\textbf{Monte-Carlo tree search with \anpi for the Tower of Hanoi puzzle.} \textbf{1}. Each simulation traverses the tree by finding the actions that maximize the sum of the action value $Q$,  
an upper confidence bound $U$ and a term $L$ that encourages programs to call programs near the same level. \textbf{2.} When the selected program is not atomic and the node has never been visited before, a new {sub-tree} is constructed. In the sub-tree, the \lstm internal state is initialized to zero.
When the sub-tree search terminates, the \lstm internal state is reset to its previous calling state. \textbf{3.} The leaf node is expanded and the associated observation $e$ and program index $i$ are evaluated by the \anpi network to compute action probabilities $P=\pi$ and values $V$. \textbf{4.} The quantities $Q$ and $U$ are computed using the network predictions. \textbf{5.} Once the search is complete, the tree policy vector $\pi^{mcts}$ is returned. 
The next program in the execution trace is chosen according to $\pi^{mcts}$, until the program \stopnpi is chosen or a computational budget is exceeded. 
}
\label{fig:npi_tree}
\end{figure}

To generate data to train the \anpi network by RL, we introduce a variant of AlphaZero using recursive \mcts. The general training procedure is illustrated in Figure~\ref{fig:npi_trace_generation}, which is inspired by Figure~1 of \cite{silver2017mastering}, but for a single-agent with hierarchical structure in this case. The Monte Carlo tree search (MCTS) guided by the AlphaNPI network enables the agent to ``imagine'' likely future scenarios and hence output an improved policy $\pi^{mcts}$, from which the next action is chosen\footnote{A detailed description of \anpi is provided in Appendix~\ref{sec:app_AlphaNPI}.}. This is repeated throughout the episode until the agent outputs the termination command \stopnpi. If the program's post-conditions are satisfied, the agent obtains a final reward of 1, and 0 otherwise. 

The data generated during these episodes is in turn used to retrain the AlphaNPI network. In particular, we record the sequence of observations, tree policies, \lstm internal states and rewards. We store the experience tuples $(e, i, h, \pi^{mcts}, r)$ in a replay buffer. The data in this replay buffer is used to train the \anpi network, as illustrated in Figure~\ref{fig:npi_trace_generation}.

The search approach is depicted in Figure~\ref{fig:npi_tree} for a Tower of Hanoi example, see also the corresponding Figure~2 of \cite{silver2017mastering}. A detailed description of the search process, including pseudo-code, appears in Appendix A. Subsequently, we present an overview of this component of \anpi.

For a specific program indexed by $i$, a node in the search tree corresponds to an observation $e$ and an edge corresponds to an action $a$. As in AlphaZero, the neural network outputs the action probabilities and node values. These values are used, in conjunction with visit counts, to compute upper confidence bounds $U$ and action-value functions $Q$ during search. Unlike AlphaZero, we add terms $L$ in the node selection stage to encourage programs not to call programs at a much lower level. In addition, we use a different estimate of the action-value function that better matches the environments considered in this paper. Actions are selected by maximizing $Q+V+L$.    

Also unlike AlphaZero, if the selected action is not atomic but an already learned program, we recursively build a new Monte Carlo tree for that program. To select a trajectory in the tree, that is the program's imagined execution trace, we play $n_{simu}$ simulations and record the number of visits to each node. This enables us to compute a tree policy vector $\pi^{mcts}$ for each node.  

The major feature of \anpi is its ability to construct recursively a new tree during the search to execute an already learned program. This approach enables to use learned skills as if they were atomic actions. When a tree is initialized to execute a new program, the \lstm internal state is initialized to zero and the environment reward signal changes to reflect the specification of the new program. The root node of the new tree corresponds to the current state of the environment. When the search process terminates, we check that the final environment state satisfies the program's post-conditions. If unsatisfied, we discard the full execution trace and start again. When returning control to an upper-level program, we assign to the \lstm the previous internal state for that level and continue the search process.

We found that discarding execution traces for programs executed incorrectly is necessary to achieve stable training. Indeed, the algorithm might choose the correct sequence of actions but still fail because one of the chosen sub-programs did not execute correctly. At the level we are trying to learn, possibly no mistake has been made, so it is wise to discard this data for training stability.

Finally, we use AlphaNPI MCTS in two different modes. In exploration mode, we use a high budget of simulations, final actions are taken by sampling according to the tree policy vectors and we add Dirichlet noise to the network priors for better exploration. This mode is used during training. In Exploitation mode, we use a low budget of simulations, final actions are taken according to the tree policy vectors argmax and we do not add noise to the priors. In this mode, \anpi's behavior is deterministic. This mode is used during validation and test.

\subsection{Training procedure}
During a training iteration, the agent selects a program $i$ to learn. It plays $n_{ep}$ episodes (See Appendix E for specific values) using the tree search in exploration mode with a large budget of simulations. The generated experiences, $(e, i, h, \pi^{mcts}, r)$, where $r$ is the episode final reward, are stored in a replay buffer. The agent is trained with the Adam optimizer on this data, so as 
to minimize the loss function:
\begin{equation}
    \label{eq:loss_fct}
    \ell = \sum_{\text{batch}} - \underbrace{\left( \pi^{mcts} \right)^T \log{ \pi }}_{\ell_{\text{policy}}} + \underbrace{(V - r)^2}_{\ell_{\text{value}}}.
\end{equation}
Note that the elements of a mini-batch may correspond to different tasks and are not necessarily adjacent in time. Given that the buffer memory is short, we make the assumption that the \lstm internal states have not changed too much. Thus, we do not use backpropagation through time to train the \lstm. Standard backpropagation is used instead, which facilitates parallelization.

After each Adam update, we perform validation on all tasks for $n_{val}$ episodes. The agent average performance is recorded and used for curriculum learning, as discussed in the following subsection. 

\subsection{Curriculum learning}
As with previous \npi models, curriculum learning plays an essential role. As programs are organized into levels, we begin by training the agent on programs of level $1$ and then increase the level when the agent's performance is higher than a specific threshold. Our curriculum strategy is similar to the one by \cite{andreas2017modular}.

At each training iteration, the agent must choose the next program to learn. We initially assign equal probability to all level 1 programs and zero probability to all other programs. At each iteration, we update the probabilities according to the agent's validation performance. We increase the probability of programs on which the agent performed poorly and decrease the probabilities of those on which the agent performed well. We compute scores $c_i=1 - R_{i}$, for each program indexed by $i$, where $R_{i}$ is a moving average of the reward accrued by this program during validation. The program selection probability is then defined as a softmax over these scores. When $\min\limits_i R_{i}$ becomes greater than some threshold $\Delta_{\text{curr}}$, we increase the maximum program level, thus allowing the agent to learn level 2 programs, and so on until it has learned every program.

\section{Experiments}
In the following experiments, we aim to assess the ability of our RL agent, \anpi, to perform the sorting tasks studied by \cite{npi} and
 \cite{recursivenpi}. We also consider a simple recursive Tower of Hanoi Puzzle. An important question we would like to answer is: Can \anpi, which is trained by RL only, perform as well as the iterarive and recursive variants of \npi, which are trained with a strong supervisory signal consisting of full execution traces? Also, how essential is MCTS planning when deploying the neural network policies?

\subsection{Sorting example}
We consider an environment consisting of a list of $n$ integers and two pointers referencing its elements. The agent can move both pointers and swap elements at the pointer positions. The goal is to learn a hierarchy of programs and to compose them to realize the \bubblesort algorithm. The library of programs is summarized in Table~\ref{table:programs_bubblesort} of the Appendix.

We trained \anpi to learn the sorting library of programs on lists of length $2$ to $7$. 
Each iteration involves 20 episodes, so the agent can see up to 20 different training lists. As soon as the agent succeeds, training is stopped, so the agent typically sees less than 20 examples per iteration.

We validated on lists of length $7$ and stopped when the minimum averaged validation reward, among all programs, reached $\Delta_{\text{curr}}$. After training, we measured the generalization of \anpi, in exploitation mode, on test lists of length 10 to 100, as shown in Table~\ref{table:sorting_generalization}. For each length, we test on 40 randomly generated lists.

\begin{table}[t!]
\centering
\begin{tabular}{c|c|c|c|c}
\hline
\multirow{2}{*}{\textbf{Length}} & \multicolumn{2}{c|}{\textbf{Iterative \bubblesort}} & \multicolumn{2}{c}{\textbf{Recursive \bubblesort}} \\
    \hhline{~----}
    & Net with planning & {Net only} & Net with planning & {Net only} \\
\hline
10 & 100\% & 85\% & 100\% & 70\% \\
20 & 100\% & 85\% & 100\% & 60\% \\
60 & 95\% & 40\% & 100\% & 35\% \\
100 & 40\% & 10\% & 100\% & 10\% \\
\end{tabular}
\vspace{1mm}
\caption{\small Performance of \anpi, trained on \bubblesort instances of length up to 7, on much longer input lists. For each \bubblesort variant, iterative and recursive, we deployed the trained AlphaNPI networks with and without MCTS planning. The results clearly highlight the importance of planning at deployment time.}
\label{table:sorting_generalization}
\end{table}

We observe that \anpi can learn the iterative \bubblesort algorithm on lists of length up to 7 and generalize well to much longer lists. The original \npi, applied to iterative \bubblesort, had to be trained with strong supervision on lists of length 20 to achieve the same generalization. As reported by~\cite{cai2017making}, when training on arrays of length 2, the iterative NPI with strong supervision fails to generalize but the recursive \npi generalizes perfectly. However, when training the recursive \npi with policy gradients RL and curricula, \cite{xiao2018improving} reports poor results.   

To assess the contribution of adding a hierarchy to the model, we trained AlphaNPI with atomic actions only to learn iterative \bubblesort. As reported on Table~\ref{table:sorting_no_hierarchy_results}, this ablation performs poorly in comparison to the hierarchical solutions.   

\begin{table}[t]
\centering
\begin{tabular}{c|c|c}
\hline
\multirow{2}{*}{\textbf{Length}} & \multicolumn{2}{c}{\textbf{Sorting without a hierarchy}} \\
    \cline{2-3}
    & Net with planning & {Net only} \\
\hline
3 & 94\% & 78\% \\
4 & 42\% & 22\%\\
5 & 10\% & 5\%\\
6 & 1\% & 1\%\\
\end{tabular}
\vspace{1mm}
\caption{\small Test performance on iterative sorting with no use of hierarchy. The AlphaNPI network is trained to sort using only atomic actions on lists of length up to 4, and tested on lists of length up to 6. The training time without hierarchy scales quadratically with list length, but only linearly with list length when a hierarchy is defined.}
\label{table:sorting_no_hierarchy_results}
\end{table}


We also defined a sorting environment in which the programs \reset, \bubble and \bubblesort are recursive. This setting corresponds to the ``full recursive'' case of \cite{recursivenpi}. Being able to learn recursive programs requires adapting environment. For instance, when a new task (recursive program) is started, the sorting environment becomes a sub-list of the original list. When the task terminates, the environment is reset to the previous list.

We trained the full recursive \bubblesort on lists of length 2 to 4 and validated on lists of length 7. After training, we assessed the generalization capabilities of the recursive AlphaNPI in Table~\ref{table:sorting_generalization}. The results indicate that the recursive version outperforms the iterative one, confirming the results reported by \cite{recursivenpi}. We also observe that AlphaNPI with planning is able to match the generalization performance of the recursive NPI with strong supervision, but that removing planning from deployment (i.e. using a network policy only) reduces performance.

\subsection{Tower of Hanoi puzzle}
We trained \anpi to solve the Tower of Hanoi puzzle recursively. Specifically, we consider an environment with $3$ pillars and $n$ disks of increasing disk size. Each pillar is given one of three roles: source, auxiliary or target. Initially, the $n$ disks are placed on the source pillar. The goal is to move all disks to the target pillar, never placing a disk on a smaller one. It can be proven that the minimum number of moves is $2^{n} - 1$, which results in a highly combinatorial problem. Moreover, the iterative solution depends on the parity of the number of disks, which makes it very hard to learn a general iterative solution with a neural network.

 To solve this problem recursively, one must be able to call the {\sc TowerOfHanoi} program to move $n-1$ disks from the source pillar to the auxiliary pillar, then move the larger disk from the source pillar to target pillar and finally call again the {\sc TowerOfHanoi} program to move the $n-1$ pillars from the auxiliary pillar to the target.

We trained our algorithm to learn the recursive solution on problem instances with 2 disks, stopping when the minimum of the validation average rewards reached $\Delta_{\text{curr}}$. Test results are shown in Table~\ref{table:hanoi_generalization}. \anpi  generalizes to instances with a greater number of disks.

In Appendix~\ref{sec:toh_proof}, we show that once trained, an \anpi agent can generalize to Tower of Hanoi puzzles with an arbitrary number of disks.

\begin{table}[t!]
\centering
\begin{tabular}{c|c|c}
\textbf{Number of disks} & \mcts & \textbf{Network only} \\
\hline
2 & 100\% & 100\% \\
5 & 100\% & 100\%\\
10 & 100\% & 100\%\\
12 & 100\% & 100\%\\
\end{tabular}
\vspace{1mm}
\caption{\small Test performance of one \anpi trained agent on the recursive Tower of Hanoi puzzle.}
\label{table:hanoi_generalization}
\end{table}

\section{Related work}



AlphaZero \citep{silver2017mastering} used Monte Carlo Tree Search for planning and to derive a policy improvement operator to train state-of-the-art neural network agents for playing Go, Chess and Shogi using deep reinforcement learning.
In \citep{rankedrewards}, AlphaZero is adapted to the setting of one-player games applied to the combinatorial problem of bin packing.
This work casts program induction as a one player game and further adapts AlphaZero to incorporate compositional structure into the learned programs.

Many existing approaches to neural program induction do not explicitly learn programs in symbolic form, but rather implicitly in the network weights and then directly predict correct outputs given query inputs. 
For example, the Neural GPU \citep{kaiser2015neural} can learn addition and multiplication of binary numbers from examples.
Neural module networks \citep{andreas2016neural} add more structure by learning  to stitch together differentiable neural network modules to solve question answering tasks.
Neural program meta induction \citep{devlin2017neural} shows how to learn implicit neural programs in a few-shot learning setting.

Another class of neural program induction methods takes the opposite approach of explicitly synthesizing programs in symbolic form.
DeepCoder \citep{balog2016deepcoder} and RobustFill \citep{devlin2017robustfill} learn in a supervised manner to generate programs for list and string manipulation using domain specific languages.
In \citep{evans2018learning}, explanatory rules are learned from noisy data.
\cite{ellis2018learning} shows how to generate graphics programs to reproduce hand drawn images.
In \citep{sun2018neural}, programs are generated from visual demonstrations.
\cite{chen2017towards} shows how to learn parsing programs from examples and their parse trees.
\cite{verma2018programmatically} shows how to distill programmatically-interpretable agents from conventional Deep RL agents.

Some approaches lie in between fully explicit and implicit, for example by making execution differentiable in order to learn parts of programs or to optimize programs \citep{bovsnjak2017programming, bunel2016adaptive, gaunt2016terpret}. In \citep{nye2019learning}, an \lstm generator conditioned on specifications is used to produce schematic outlines of programs, which are then fed to a simple logical program synthesizer. Similarly, \cite{shin2018improving} use \lstms to map input-output pairs to traces and subsequently map these traces to code.

Neural Programmer-Interpreters \citep{npi}, which we extend in this work, learn to execute a hierarchy of programs from demonstration.
\cite{recursivenpi} showed that by learning recursive instead of iterative forms of algorithms like bubble sort, \npi can achieve perfect generalization from far fewer demonstrations. Here, perfect generalization means generalization with provable theoretical guarantees.
Neural Task Programming \citep{xu2018neural} adapted \npi to the setting of robotics in order to learn manipulation behaviors from visual demonstrations and annotations of the program hierarchy.

Several recent works have reduced the training data requirements of \npi, especially the ``strong supervision'' of demonstrations at each level of the program hierarchy.
For example, \cite{li2016neural} and \cite{fox2018parametrized} show how to train variations of \npi using mostly low-level demonstration trajectories and a relatively smaller proportion of hierarchical annotations compared to \npi. 
However, demonstrations are still required. 
\cite{xiao2018improving} incorporates combinator abstraction techniques from functional programming into \npi to improve training, but emphasize the difficulty of learning simple \npi models with RL algorithms.

Hierarchical reinforcement learning combined with deep neural networks has received increased attention in the past several years \citep{osa2019hierarchical, nachum2018data, kulkarni2016hierarchical, nachum2018near, levy2018hierarchical, vezhnevets2017feudal}, mainly applied to efficient training of agents for Atari, navigation and continuous control. 
This work shares a similar motivation of using hierarchy to improve generalization and sample efficiency, but we focus on algorithmic problem domains and learning potentially recursive neural programs without any demonstrations.

While AlphaZero does not use hierarchies or recursion, hierarchical MCTS algorithms have been previously proposed for simple hierarchical RL domains \citep{vien2015hierarchical,bai2016markovian}. The current work capitalizes on advances brought in by deep reinforcement learning as well as design choices particular to this paper to significantly extend this research frontier. 

Finally, as demonstrated in the original NPI paper, the modular approach with context-dependent input embeddings and a task independent interpreter is ideal for meta-learning and transfer. Recent manifestations of this idea of using an embedding to re-program a core neural network to facilitate meta-learning include \cite{zintgraf2018fast} and \cite{chen2019sample}. To the best of our knowledge the idea of programmable neural networks goes back several decades to the original Parallel Distributed Programming (PDP) papers of Jay McClelland and colleagues. We leave transfer and meta-learning as a future explorations for \anpi.

\section{Conclusion}

This paper proposed and demonstrated the first effective RL agent for training \npi models: \anpi.
\anpi extends NPI to the RL domain and enhances AlphaZero with the inductive biases of modularity, hierarchy and recursion. 
\anpi was shown to match the performance of strongly supervised versions of \npi in the sorting experiment, and to generalize remarkably well in the Tower of Hanoi environment.
The experiments also shed light on the issue of deploying neural network RL policies. Specifically, we found out that agents that harness MCTS planning at test time are much  more effective than plain neural network policies. 

While our test domains are complex along some axes, e.g. recursive and combinatorial, they are simple along others, e.g. the environment model is available. The natural next step is to consider environments, such as robot manipulation, where it is also important to learn perception modules and libraries of skills in a modular way to achieve transfer to new tasks with few data. It will be fascinating to harness imperfect environment models in these environments and assess the performance of MCTS planning when launching AlphaNPI policies.

\section*{Acknowledgements}

Work by Nicolas Perrin was partially supported by the French National Research Agency (ANR), Project ANR-18-CE33-0005 HUSKI.

\newpage

{\small
\bibliography{ms}
\bibliographystyle{plainnat}
}
\clearpage

\newpage

\appendix

\section*{Appendices}

\section{Detailed description of AlphaNPI}
\label{sec:app_AlphaNPI}

When learning the $i$-th program, the environment is reset to a state that satisfies the program pre-conditions. At the same time, we adopt a reward that returns 1 when the program post-condition is satisfied and 0 otherwise. A tree is built for this specific task, to maximize the expected reward. 

For a specific program, a node corresponds to an environment state\footnote{Strictly speaking, the node is for the pair $(e,i)$ but we are assuming a fixed $i$ and dropping the program index to simplify notation.} $e$ and an edge corresponds to an action $a$. The root node corresponds to the initial environment state. Every node contains a prior, a visit count and a Q-value estimate. The prior $P(e, a)$ corresponds to the probability of this node being chosen by its parent node. The visit count $N(e, a)$ records how many times this node has been visited through simulations and the Q-value $Q(e, a)$ estimates the expected reward the agent will accrue if it chooses action $a$ from this node. 

A simulation involves three operations: select, expand and evaluate, and value backup. When a selected action corresponds to a non-zero level program, we recursively build a new tree to execute it, see Algorithm \ref{alg:tree_simu}. Finally, when a given budget of simulations has been spent, a tree policy vector is computed and the next action is chosen according to this vector. We delve into the details of these steps in the following subsections. These steps are illustrated in Figure~\ref{fig:npi_tree}.

\subsection{Select}
\label{subsec:mcts_puct_criteria}
From a non-terminal node, the next action is chosen to maximise the P-UCT criterion:

\begin{equation}
\label{eq:puct_criteria}
    a = \argmax\limits_{a' \in \mathcal{A}} \left( Q(e, a') + U(e, a') + L(i, a') \right),
\end{equation}
\begin{equation}
\label{eq:utility_term}
U(e, a) = c_{\text{puct}}P(e, a)\dfrac{\sqrt{\sum_b N(e, b)}}{1 + N(e, a)}.
\end{equation}

The coefficient $c_{\text{puct}}$ is user-defined and trades-off exploration and exploitation. The level balancing term $L$ is defined as:
\begin{equation}
\label{eq:level_puct_term}
    \left\{
      \begin{array}{l}
        L(i, a) = c_{\text{level}} \exp{(-1)}, \ \ \ \text{if} \ a \ \text{is STOP}\\
        L(i, a) = c_{\text{level}} \exp{(-1)}, \ \ \ \text{if} \ \text{level}(i)=\text{level}(a)\\
        L(i, a) = c_{\text{level}} \exp{\left(-\left( \text{\text{level}}(i) - \text{level}(a) \right)\right) }, \ \ \ \text{otherwise}\\
      \end{array}
    \right.
\end{equation}
where $c_{\text{level}}$ is a user-defined constant and $\text{level}$ is an operator that returns a program level. This term encourages programs to call programs near them in the hierarchy.

We perform additional exploration as in the original AlphaZero work of \cite{silver2017mastering} by adding Dirichlet noise to the priors:
\begin{equation}
    \label{eq:dirichlet_noise}
    P(e,a) \longleftarrow (1-\epsilon_d)P(e,a) + \epsilon_d\eta_d, \ \ \text{where} \ \eta_d \sim \text{Dir}(\alpha_d)
\end{equation}
where $\eta_d$ follows a Dirichlet distribution with hyper-parameters $\alpha_d$.

\subsection{Tree recursion}
 If the chosen action is atomic, we apply it directly in the environment and the new environment observation is recorded inside the node. Otherwise, a new tree is built to execute the newly invoked program. In this case, the environment reward changes to correspond to this new task (program), and the \lstm internal state is re-initialized to zero. The new tree is built in exploitation mode. When the search terminates, we check if the program post-conditions are satisfied. If unsatisfied, we stop the entire search and discard the corresponding trace. If satisfied, the task at hand becomes the previous one (return to calling program). In this case, the \lstm is assigned its previous internal state and the new environment state is recorded inside the child node. From this point of view, the program has been executed as if it was an atomic action.

\clearpage
\begin{algorithm}[t]
\SetAlgorithmName{Algorithm}{}
\KwData{}
{\small
\textbf{Input}: Node $n=(e,i)$ and \lstm internal state $h$

\While{True}{
    
    \eIf{$n$ has not been expanded}{
         Compute possible children nodes to respect programs levels and pre-conditions\;
         Evaluate the node with NPI network to compute priors and V-value\;
         If the mode is exploration add Dirichlet noise to the priors\;
         Get and store new \lstm internal state $h$\;
         Store the priors in the node\;
         Stop the simulation and return V-value\;
   }{
         Select an action $a$ to according to Equation \ref{eq:puct_criteria}\;
         \eIf{simulation length $\geq$ maximum tree depth}{
                Stop the simulation \;
                Return a reward of -1
            }{
                    \eIf{$a ==$ STOP}{
                    Stop the simulation \;
                    Return the obtained reward
                 }{
                    \eIf{$a$ is a level 0 program}{
                        Apply $a$ in the environment\footnotemark\setcounter{auxFootnote}{\value{footnote}}\;
                    }
                    {
                        Build a new tree\footnotemark[\value{footnote}] in exploitation mode to execute $a$ \;
                    }
                    Record new environment observation $e^\prime$\;
                    Consider new node $n = (e^\prime, i)$\;
                 }
            
            }
   }
    
}
}
\caption{Perform one MCTS simulation}
\label{alg:tree_simu}
\end{algorithm}
\footnotetext{When an action $a$ is called, we apply it into the environment only the first time. The resulting environment state is stored inside the outgoing node. When this action is to be called again from the same node we reset the environment in the correct environment state. We apply the same strategy when $a$ corresponds to a non-atomic program. The recursive tree is built only the first time.}

\begin{algorithm}[!ht]
\SetAlgorithmName{Algorithm}{}
\KwData{}
{\small
\textbf{Input}: program index $i$ and a mode (exploration/exploitation)

Initialize \lstm internal state $h$ to 0

Initialize execution trace to empty list

Get initial environment observation $e_0$

Build root node $n = (e_0, i)$

\While{$True$}{
    
    \For{\emph{k = 1}, \dots, $n_{simu}$}{
        Reset the environment in the state corresponding to root node $n$
        
    	Perform one simulation from root node $n$ using Algorithm 1
    	
    	Back-up values in the tree
    	
    	Update nodes visit counts
    }  
    
    Compute tree policy $\pi^{mcts}$ with visit counts
    
    Choose next action $a \sim \pi^{mcts}$ (exploration) or $a = \argmax{\pi^{mcts}}$ (exploitation)
    
    Add action to the execution trace
    
    \eIf{trace length $\geq$ maximum tree depth}{
        Stop the search\;
        Return the execution trace and a -1 reward\;
    }{
        \eIf{$a == $ STOP}{
         Stop the search\;
         Get final reward\;
         Return the execution trace and the final reward\;
       }{
         From root node select the edge corresponding to $a$\;
         The outgoing node becomes the new root node $n$\;
         Reset the environment in the state corresponding to new root node $n$
       }
   
    }
}    
}
\caption{AlphaNPI tree search}
\label{alg:tree_search}
\end{algorithm}

\newpage

\subsection{Expand and evaluate}
When a node is expanded, we construct a new child node for every possible action available to the parent node. The possible actions correspond to the programs whose pre-conditions are satisfied and whose level is lower (or equal if the setup is recursive) than the current program's level. The priors over the edges and node V-value are computed using the \anpi network. The child nodes' Q-values are initialized to 0 and their visit counts to 0.

\subsection{Value back-up}
When an action is chosen, if the new node is not terminal, we compute its value with the value head of the \anpi network. Otherwise, we use the episode's final reward. When the obtained reward equals 0, we replace it by a value of -1 as the Q-values are initialized to 0. To encourage short traces, we penalize positive rewards by multiplying them by $\gamma^n$, where $n$ is the trace length and $\gamma \in [0,1]$. This value is then backpropagated on the path between the current node and the root node. The visit counts of the nodes in this path are also incremented. For each node, we maintain a list of all the values that have been backpropagated. In classical two-player \mcts, the Q-value is computed as the sum of the values, generated by the neural network for the child node, divided by the visit count. Since our approach is single-player, we use a slightly different expression:
\begin{equation}
\label{eq:qvalue_expression}
\left.
      \begin{array}{l}
Q(e, a) = \sum\limits_{e^\prime | e,a \rightarrow e^\prime} p_{e^\prime} V(e^\prime),\\
p_{e^\prime} = \dfrac{\exp{\left( \tau_1 V(e^\prime) \right)}}{\sum\limits_{e^\prime | e,a \rightarrow e^\prime} \exp{\left( \tau_1 V(e^\prime) \right)}},
\end{array}
    \right.
\end{equation}
where $\tau_1$ is a temperature coefficient and $e^\prime | e,a \rightarrow e^\prime$ indicates that a simulation eventually reached $e^\prime$ after taking action $a$ from node $e$.
In two-player \mcts, the expected reward does not depend only on the chosen action but also on the other player’s response. Due to the stochasticity (typically adversarial) of this response, it is judicious to choose actions with good Q-values on average. In our single-player approach, the environment is deterministic, therefore we focus on a highly rewarding course of actions.

\subsection{Final execution trace construction}
To compute the final execution trace, we begin at the tree root-node and launch $n_{\text{simu}}$ simulations. We choose the next action in the execution trace according to the nodes' visit counts. We compute tree policy vectors
\begin{equation}
    \pi^{mcts}(a) = \dfrac{N(e, a)^{\tau_2}}{\sum_b N(e, b)^{\tau_2}},
\end{equation}
where $\tau_2$ is a temperature coefficient. If the tree is in exploration mode, the next action is sampled according to this probability vector. If the tree is in exploitation mode, it is taken as the tree policy argmax. When an action is chosen, the new node becomes the root-node, and $n_{\text{simu}}$ simulations are played from this node and so-on until the end of the episode, see Algorithm \ref{alg:tree_search}. The final trajectory is then stored inside a replay buffer.

\subsection{Prioritized replay buffer}
The experience generated by \mcts is stored inside a  prioritized replay buffer. This experience takes the form of tuples $(e, i, h, \pi^{mcts}, r)$ where $e$ is an environment observation, $i$ the program index, $h$ the \lstm internal state, $\pi^{mcts}$ the tree policy for the corresponding node and $r$ the reward obtained at the end of the trajectory. The buffer has a maximum memory size $n_{buf}$. When the memory is full, we replace a past experience with a new one. To construct a training batch, we sample buffer tuples with probability $p_{buf}$, which measures the chance that the tuple results in positive reward. We also make sure that the buffer does not contain experiences related to tasks for which a positive reward has not been found yet.

\subsection{Network architecture}
To adapt the original \npi algorithm to our setting, me make the following modifications:
\begin{enumerate}
\item Programs do not accept arguments anymore. The program ACT and its finite set of possible arguments are replaced by atomic actions, one for each argument.
\item The termination scalar returned by the network is replaced by the action {\sc stop}.
\item For simplicity, the program keys matrix has been replaced by a dense layer with identity activation function. 
\item We added policy and value modules to the architecture to obtain an actor-critic architecture necessary for RL.
\end{enumerate}
In our architecture, the \lstm core has one layer of $H$ neurons. Both the actor and the critic modules are multi-layer perceptrons with one hidden layer of $H/2$ neurons and ReLu activation functions. The encoder is environment dependent, however in the three environments we consider, it is a simple multi-layer perceptron with one hidden layer.

\subsection{Curriculum learning}
In the curriculum scheduler, we maintain a maximum program level $l_{max}$, which is initialized to 1. At the beginning of each training iteration, the curriculum scheduler is called to determine the next program to be learned. This program must have a level lower than $l_{max}$. Each time a validation episode for the $i$-th program is conducted, we record the final reward $r$ and update the $i$-th program average reward $R_i$ as follows:
\begin{equation}
    \label{eq:curriculum_averaged_r}
    R_i \longleftarrow \beta R_i + (1-\beta) r
\end{equation}
where $\beta$ is a user-defined coefficient.

When the minimum average reward $R_i$ over the programs of level lower than $l_{max}$ reaches $\Delta_{\text{curr}}$ we increment $l_{max}$. To determine which program should be learned next, the curriculum scheduler computes probabilities over the programs of level lower than its internal maximum level. The $i$-th program probability $p_i$ is defined as follows
\begin{equation}
    \label{eq:curriculum_proba}
\left.
      \begin{array}{l}
    p_i = \dfrac{\exp{(\tau_3 c_i)}}{\sum\limits_k \exp{( \tau_3 c_k)}}\\
    c_i = \nicefrac{1}{R_i}
\end{array}
    \right.
\end{equation}
where $\tau_3$ is a temperature coefficient.

\section{Environments}
\subsection{Sorting environment}

We consider a list of $n$ digits and two pointers that refer to elements in the list. The atomic actions are moving the pointers and swapping elements at the pointer locations. The level 1 programs may move both pointers at the same time and conditionally swap elements. Level 2 programs are {\sc reset} and {\sc bubble}. {\sc reset} moves both pointers to the extreme left of the list and {\sc bubble} conditionally compares two by two the elements from left to right. \bubblesort is level 3 and sorts the list.

In this environment, compositionality is mandatory for the tree search to find a way to sort the list when $n$ is greater than 3. Indeed, \bubble requires $3n$ atomic actions and \reset $2n$ atomic actions. When both programs are known, \bubblesort simply alternates \bubble and \reset $n$ times. Therefore, if \bubblesort had to use atomic actions only, it would require $n\times 3n + n\times 2n = 5n^2$ actions, while it might require only a correct sequence of $2n$ actions if \bubble and \reset programs have already been learned.

The environment observations have the form $e=(v_1, v_2, b_{1i}, b_{1e}, b_{2i}, b_{2e}, b_{12}, b_s)$
where $v_1$ and $v_2$ are the one-hot-encoded vectors that represent the digits referenced by the pointers 1 and 2. $b_1i$, $b_2i$ and $b_1e$, $b_2e$ respectively equal 1 if the pointer 1/2 is at the beginning/end of the list and 0 otherwise. $b_{12}$ equals 1 if both pointers are at the same position and 0 otherwise. $b_s$ equals 1 if the list is sorted and 0 otherwise.
The dimension of the observation space is 26.
The encoder is composed of one hidden layer with 100 neurons and a ReLu activation function. 

The program library specification appears in Table~\ref{table:programs_bubblesort}, with the  pre-conditions defined in Table~\ref{table:programs_bubblesort2}.
\begin{table}[t!]
\centering
\begin{tabular}{|l|l|c|}
\hline
\textbf{program} & \textbf{description} & \textbf{level} \\
\hline
\hline
{\sc Bubblesort} & sort the list & 3\\
\hline
{\sc Reset} & move both pointers to the extreme left of the list & 2\\
Bubble & make one pass through the list & 2\\
\hline
{\sc Rshift} & move both pointers once to the right & 1\\
{\sc Lshift} & move both pointers once to the left & 1\\
{\sc Compswap} & \Longunderstack[l]{\noindent if both pointers are at the same position, move pointer 2 to the left,\\ then swap elements at pointers positions if left element > right element} & 1\\
\hline
{\sc Ptr\_2\_l} & move pointer 2 to the left & 0\\
{\sc Ptr\_1\_l} & move pointer 1 to the left & 0\\
{\sc Ptr\_1\_r} & move pointer 1 to the right & 0\\
{\sc Ptr\_2\_r} & move pointer 2 to the right & 0\\
{\sc Swap} & swap elements at the pointers positions & 0\\
{\sc Stop} & terminates current program & 0\\
\hline
\end{tabular}
\vspace{1mm}
\caption{Program library for the list sorting environment.}
\label{table:programs_bubblesort}
\end{table}
\begin{table}[t]
\centering
\begin{tabular}{|l|l|c|}
\hline
\textbf{program} & \textbf{pre-condition} \\
\hline
\hline
{\sc Bubblesort} & both pointers are at the extreme left of the list\\
\hline
{\sc Reset} & both pointers are not at the extreme left of the list\\
{\sc Bubble} & both pointers are at the extreme left of the list\\
\hline
{\sc Rshift} & both pointers are not at the extreme right of the list\\
{\sc Lshift} & both pointers are not at the extreme left of the list\\
{\sc Compswap} & pointer 1 is directly at the left of pointer 2, or they are at the same position\\
\hline
{\sc Ptr\_2\_l} & pointer 2 is not at the extreme left of the list\\
{\sc Ptr\_1\_l} & pointer 1 is not at the extreme left of the list\\
{\sc Ptr\_1\_r} & pointer 1 is not at the extreme right of the list\\
{\sc Ptr\_2\_r} & pointer 2 is not at the extreme right of the list\\
{\sc Swap} & the pointers are not at the same position\\
{\sc Stop} & no condition\\
\hline
\end{tabular}
\vspace{1mm}
\caption{Program pre-conditions for the list sorting environment.}
\label{table:programs_bubblesort2}
\end{table}

\subsection{Recursive Sorting environment}
We consider the same environment and the same programs library than for the non-recursive case. The only difference is the environment ability to decrease the size of the list when a task corresponding to recursive program starts and to increase back its size when the task ends.\\

Learning recursive programs in this environment has the strong advantage to remove the dependency to execution traces length. Indeed, in the non-recursive case, the size of the execution traces of Reset, Bubble and Bubblesort depends linearly of the list length. Their execution traces size are constant in the recursive case which facilitates the tree search.

\subsection{Tower of Hanoi environment}

We consider three pillars and $n$ disks. When a game is started, each pillar is attributed an initial role that can be source, auxiliary or target. The $n$ disks are initially placed on the source pillar in decreasing order, the largest one being at the bottom. The goal is to move all disks from the source pillar to the target pillar, without ever placing a disk on a smaller one. For each pillar, we consider its initial role and its current role. At the beginning, both are equivalent. Acting consists of switching the current roles of two pillars and moving a disk from the current source pillar to the current target pillar.

The game starts when the program {\sc TowerOfHanoi} is called. If during a game the {\sc TowerOfHanoi}  program is called again, i.e. is called recursively, the largest disk is removed and the game restarts. The roles of the initial pillars become the current roles in the previous game. The reward signal changes accordingly. When {\sc TowerOfHanoi} terminates, the largest disk is placed back at its previous location and the pillars get the initial roles they had in the previous game.

\begin{figure}[htbp]
\label{fig:hanoi_env}
\centering
\includegraphics[width=7cm]{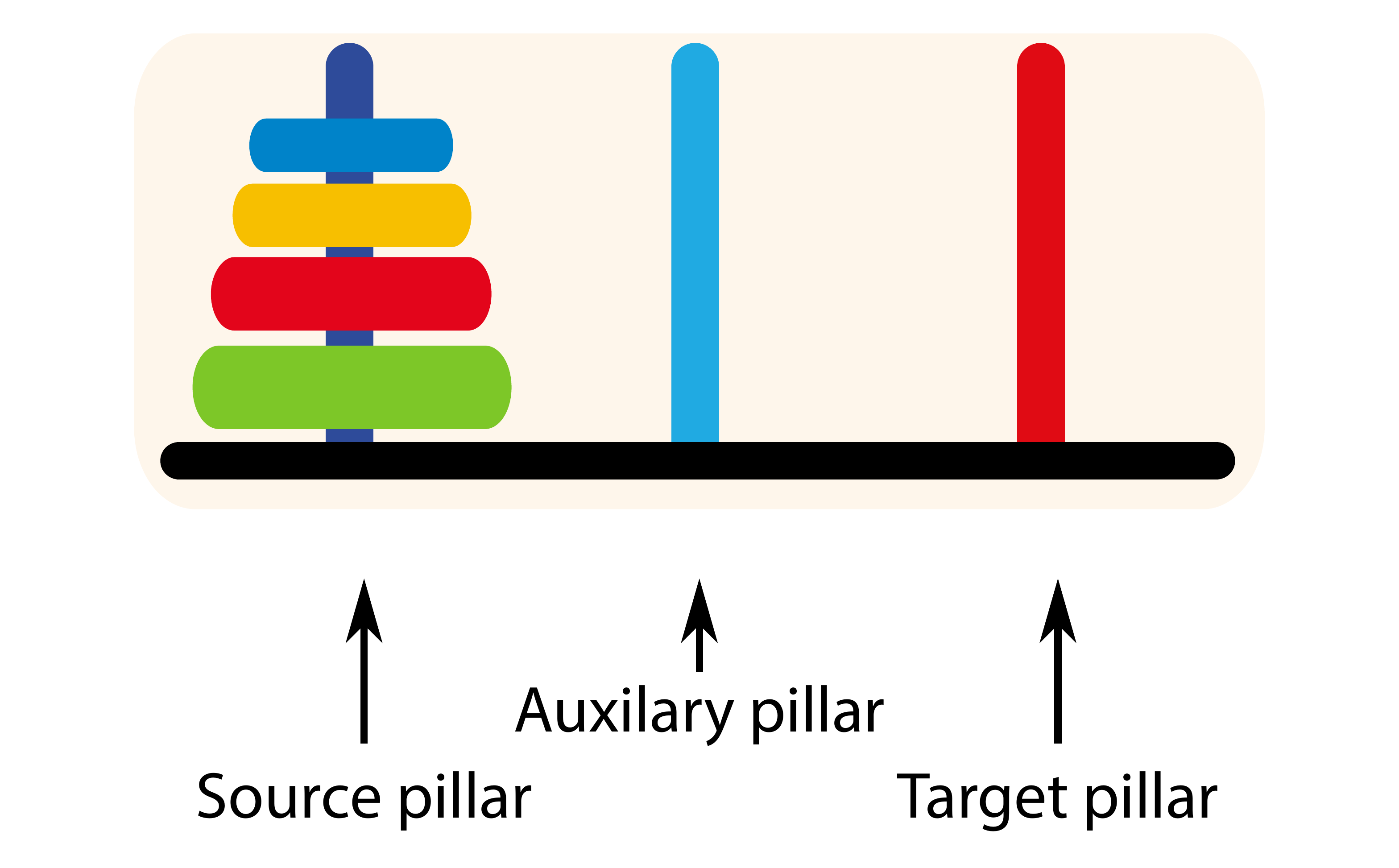}
\caption{Tower of Hanoi environment illustration.}
\end{figure}

The combinatorial nature of the Tower of Hanoi puzzle, and in particular its sparse reward signal, makes this game a challenge for conventional reinforcement learning algorithms. In \citep{ForwardBackwardHanoi}, the authors introduced a backward induction, to enable the agent to reason backwards in time. Through an iterative process, it both explores forwards from the start position and backwards from the target/goal. They have shown that by endowing the agent with knowledge of the reward function, and in particular of the goal, it can outperform the standard DDQN algorithm. However, their experiments were limited to three-disks which they solved perfectly without mentioning any generalisation performance beyond this number.


The environment observations have the form $e=(m_1, m_2, m_3, b_n, b_s)$ where $m_1$, $m_2$ and $m_3$ are equal to 1 if respectively the move from the source/auxiliary/source pillar 
to the auxiliary/target/target pillar is possible, and 0 otherwise.
$b_n$ equals 1 if $n=1$ and 0 otherwise.
$b_s$ equals 1 if the puzzle is solved, i.e. all the disks are on the 
target pillar and the target pillar is in its initial location.Therefore, the observations dimension is 5. 
The encoder is composed of one hidden layer with 100 neurons and a ReLu activation function.


The program library specification appears in Table~\ref{table:programs_hanoi}, while the
 pre-conditions are defined in Table~\ref{table:programs_hanoi2}.
\begin{table}[ht!]
\centering
\begin{tabular}{|l|l|c|}
\hline
\textbf{program} & \textbf{description} & \textbf{level} \\
\hline
\hline
{\sc TowerOfHanoi} & move $n$ disks from source pillar to target pillar & 1\\
\hline
{\sc Swap\_s\_a} & source pillar becomes auxiliary and vice-versa & 0\\
{\sc Swap\_a\_t} & auxiliary pillar becomes target and vice-versa & 0\\
{\sc MoveDisk} & move disk from source to target & 0\\
{\sc Stop} & terminates current program & 0\\
\hline
\end{tabular}
\vspace{1mm}
\caption{Program library for Tower of Hanoi.}
\label{table:programs_hanoi}
\end{table}
\begin{table}[ht!]
\centering
\begin{tabular}{|l|l|c|}
\hline
\textbf{program} & \textbf{pre-condition} \\
\hline
\hline
{\sc TowerOfHanoi} & all $n$ disks are on the source pillar\\
\hline
{\sc Swap\_s\_a} & the number of disks is greater than one\\
{\sc Swap\_a\_t} & the number of disks is greater than one\\
{\sc MoveDisk} & the move from source to target is possible\\
{\sc Stop} & no pre-condition\\
\hline
\end{tabular}
\vspace{1mm}
\caption{Program pre-conditions for Tower of Hanoi.}
\label{table:programs_hanoi2}
\end{table}

The {\sc TowerOfHanoi} post-condition is satisfied when $n$ disks have been moved from the initial source pillar to the initial target pillar and all pillars' current roles correspond to their initial roles. When {\sc TowerOfHanoi} is called recursively, its 
pre-condition is tested in the new environment with the largest disk removed.

\section{Tower of Hanoi recursion proof}
\label{sec:toh_proof}

In this section, we prove that once trained \anpi can generalize to Hanoi puzzles with an arbitrary number of disks.

We remind the reader that the environment observations have the form $e=(m_1, m_2, m_3, b_n, b_s)$ where $m_1$, $m_2$ and $m_3$ are equal to 1 if respectively the move from the source/auxiliary/source pillar 
to the auxiliary/target/target pillar is possible, and 0 otherwise.
$b_n$ is equal to 1 if $n=1$ and 0 otherwise.
$b_s$ is equal to 1 if the puzzle is solved, i.e. all the disks are on the target pillar and all pillars are at their initial locations. Otherwise, $b_s=0$.

The environment is initialized with all disks on the source pillar. Therefore, there are only two possible initial observations:
$e_0^1 = (1,0,1,1,0)$ if there is only 1 disk, and $e_0^n = (1,0,1,0,0)$ if $n\geq2$.

In exploitation mode, \anpi has a deterministic behavior, so two identical sequences of observations necessarily correspond to the exact same sequence of actions.

We assume that the trained agent solves the case $n=1$, and that, for $n=2$, it solves the Hanoi puzzle with the following sequence of observations and actions: 
\begin{itemize}
    \item[--] $e_0^2=(1, 0, 1, 0, 0)$ $\rightarrow$ {\sc Swap\_a\_t}
    \item[--] $e_1^2=(1, 0, 1, 0, 0)$ $\rightarrow$ {\sc TowerOfHanoi}
    \item[--] $e_2^2=(1, 0, 0, 0, 0)$ $\rightarrow$ {\sc Swap\_a\_t}
    \item[--] $e_3^2=(0, 1, 1, 0, 0)$ $\rightarrow$ {\sc MoveDisk}
    \item[--] $e_4^2=(0, 1, 0, 0, 0)$ $\rightarrow$ {\sc Swap\_s\_a}
    \item[--] $e_5^2=(1, 0, 1, 0, 0)$ $\rightarrow$ {\sc TowerOfHanoi}
    \item[--] $e_6^2=(0, 0, 0, 0, 1)$ $\rightarrow$ {\sc Swap\_s\_a}
    \item[--] $e_7^2=(0, 0, 0, 0, 1)$ $\rightarrow$ {\sc Stop}
\end{itemize}

For any $n\geq3$, the initial observation is the same: $e_0^n = e_0^2 =  (1,0,1,0,0)$, leading to the same action {\sc Swap\_a\_t}. The next
observation is again $(1,0,1,0,0)$, so the second action is also {\sc TowerOfHanoi}. Assuming that the recursive call to {\sc TowerOfHanoi} is successful (i.e. the case $n-1$ is solved), it can be verified that the exact same sequence of 8 observations and actions is generated. Besides, if the recursive call to {\sc TowerOfHanoi} is successful, this sequence of 8 actions actually solves the puzzle. By induction, we conclude that for any $n\geq2$, the agent generates the same sequence of actions, which solves the puzzle.

This proof shows that by simply observing the behavior of the trained agent on the cases with 1 and 2 disks, we can possibly acquire the certainty that the agent generalizes correctly to any number of disks.
By encouraging agents to try recursive calls during their training (see Section~\ref{sec:rec_programs}), AlphaNPI agents often end up solving the case $n=2$ with the above sequence of actions. So, even though this generalization proof does not apply to every trained agent, it is often a convenient way to verify the correctness of the agent's behavior for any number of disks.

\section{Implementation details}
The code has been developed in Python3.6. We used Pytorch as the Deep Learning library. The code is not based on existing implementations. Our \anpi architecture, the search algorithm and the environments have been developed from scratch. The code is open-source.

\subsection{Recursive programs}
\label{sec:rec_programs}
When specifying a program library, the user can define recursive programs. In this setting, a program can call lower level programs and itself. When the program calls itself, a new tree is recursively built to execute it and everything happens as for any other program call. As we train the recursive programs on small problem instances, for example on small lists, \mcts is likely to find non-recursive solutions. Hence, when a program is defined as recursive by the user, we encourage the algorithm to find a recursive execution trace, i.e. an execution in which the program calls itself, by subtracting from the reward of non-recursive programs a constant $r_{pen-recur}$. Note that this helps the algorithm find recursive solutions, but does not completely prevent it from finding non-recursive ones.

\subsection{Computation resources}
We trained \anpi on the three environments with a 12 CPUs laptop with no GPU. With this computing architecture, training on one environment (Tower of Hanoi or \bubblesort) takes approximately 2 hours.


\section{Hyper-parameters}

\begin{table}[ht!]
\centering
\begin{tabular}{|l|l|r|}
\hline
\textbf{Notation} & \textbf{description} & \textbf{value} \\
\hline
\hline
$P$ & program embedding dimension & 256\\
$H$ & \lstm hidden state dimension & 128\\
$S$ & observation encoding dimension & 32\\
$\Delta_{curr}$ & threshold in curriculum learning & 0.97\\
$\gamma$ & discount factor to penalize long traces reward & 0.97\\
$n_{simu}$ & number of simulations performed in the tree in exploration mode & 200/1500\footnotemark\setcounter{auxFootnote}{\value{footnote}} \\
$n_{simu-exploit}$ & number of simulations performed in the tree in exploitation mode & 5\\
$n_{batches}$ & number of batches used for training at each iteration & 2\\
$n_{ep}$ & number of episodes played at each iteration & 20\\
$n_{val}$ & number of episodes played for validation & 25\\
$c_{\text{level}}$ & coefficient to encourage choice of higher level programs & 3.0\\
$c_{\text{puct}}$ & coefficient to balance exploration/exploitation in \mcts & 0.5\\
{\it batch size} & batch size & 256\\
$n_{buf}$ & buffer memory maximum size & 2000\\
$p_{buf}$ & probability to draw positive reward experience in buffer & 0.5\\
$lr$ & learning rate & 0.0001\\
$r_{pen-recur}$ & penalty to encourage recursive execution trace & 0.9\\
$\tau_1$ & q-values computation temperature coefficient & 1.0\\
$\tau_2$ & tree policies temperature coefficient & 1.3\\
$\tau_3$ & curriculum temperature coefficient & 2.0\\
$\beta$ & curriculum scheduler moving average & 0.99\\
$\epsilon_d$ & AlphaZero Dirichlet noise fraction & 0.25/0.5\footnotemark[\value{footnote}]\\
$\alpha_d$ & AlphaZero Dirichlet distribution parameter & 0.03/0.5\footnotemark[\value{footnote}]\\
\hline
\end{tabular}
\vspace{1mm}
\caption{Hyperparameters.}
\label{table:hyperparameters_table}
\end{table}
\footnotetext{We use a greater number of simulations and add more Dirichlet noise to train \anpi on the Tower of Hanoi than to train it on \bubblesort because of the higher complexity of the problem.}

\end{document}